\documentclass[11pt,a4paper]{article}
\usepackage[latin9]{inputenc}
\usepackage{array}
\usepackage{rotating}
\usepackage{multirow}
\usepackage{amstext}
\usepackage{graphicx}
\usepackage{setspace}
\usepackage[authoryear]{natbib}
\usepackage[unicode=true,pdfusetitle,
 bookmarks=true,bookmarksnumbered=false,bookmarksopen=false,
 breaklinks=false,pdfborder={0 0 1},backref=false,colorlinks=false]
 {hyperref}

\makeatletter

\pdfpageheight\paperheight
\pdfpagewidth\paperwidth

\providecommand{\tabularnewline}{\\}

\@ifundefined{date}{}{\date{}}
\usepackage[hyperref]{naaclhlt2018}
\usepackage{times}
\usepackage{latexsym}
\usepackage[font=small,labelfont=bf]{caption}

\usepackage{url}

\aclfinalcopy 

\renewcommand{\and}{\end{tabular}\kern-\tabcolsep\ and\ \kern-\tabcolsep\begin{tabular}[t]{c}}

\title{Amobee at SemEval-2018 Task 1: GRU Neural Network with a CNN Attention Mechanism for Sentiment Classification}

\author{Alon Rozental\thanks{~~These authors contributed equally to this work.}~~, Daniel Fleischer\footnotemark[1] \\
  Amobee, Tel Aviv, Israel \\
   \tt alon.rozental@amobee.com \\ \tt daniel.fleischer@amobee.com }

\makeatother

\begin{document}
\maketitle
\begin{abstract}
This paper describes the participation of Amobee in the shared sentiment
analysis task at SemEval 2018. We participated in all the English
sub-tasks and the Spanish valence tasks. Our system consists of three
parts: training task-specific word embeddings, training a model consisting
of gated-recurrent-units (GRU) with a convolution neural network (CNN)
attention mechanism and training stacking-based ensembles for each
of the sub-tasks. Our algorithm reached 3rd and 1st places in the
valence ordinal classification sub-tasks in English and Spanish, respectively. 
\end{abstract}

\section{Introduction\label{sec:Introduction}}

Sentiment analysis is a collection of methods and algorithms used
to infer and measure affection expressed by a writer. The main motivation
is enabling computers to better understand human language, particularly
sentiment carried by the speaker. Among the popular sources of textual
data for NLP is Twitter, a social network service where users communicate
by posting short messages, no longer than 280 characters long\textemdash called
tweets. Tweets can carry sentimental information when talking about
events, public figures, brands or products. Unique linguistic features,
such as the use of slang, emojis, misspelling and sarcasm, make Twitter
a challenging source for NLP research, attracting the interest of
both academia and the industry. 

Semeval is a yearly event in which international teams of researchers
work on tasks in a competition format where they tackle open research
questions in the field of semantic analysis. We participated in Semeval
2018 task 1, which focuses on sentiment and emotions evaluation in
tweets. There were three main problems: identifying the presence of
a given emotion in a tweet (sub-tasks EI-reg, EI-oc), identifying
the general sentiment (valence) in a tweet (sub-tasks V-reg, V-oc)
and identifying which emotions are expressed in a tweet (sub-task
E-c). For a complete description of Semeval 2018 task 1, see the official
task description \citep{SemEval2018Task1}.

We developed an architecture based on gated-recurrent-units (GRU,
\citet{Cho:2014aa}). We used a bi-directional GRU layer, together
with a convolutional neural network (CNN) attention-mechanism, where
its input is the hidden states of the GRU layer; lastly there were
two fully connected layers. We will refer to this architecture as
the Amobee sentiment classifier (ASC). We used ASC to train word embeddings
to incorporate sentiment information and to classify sentiment using
annotated tweets. We participated in all the English sub-tasks and
in the valence Spanish sub-tasks, achieving competitive results. 

The paper is organized as follows: section \ref{sec:Data} describes
our data sources, section \ref{sec:Data-Cleanup} describes the data
pre-processing pipeline. A description of the main architecture is
in section \ref{sec:RNN-Models}. Section \ref{sec:Word-Embeddings}
describes the word embeddings generation; section \ref{sec:Features-Extraction}
describes the extraction of features. In section \ref{sec:Experiments}
we describe the performance of our models; finally, in section \ref{sec:Review-and-Conclusions}
we review and summarize the results.

\section{Data Sources\label{sec:Data}}

We used four sources of data: 
\begin{enumerate}
\item \setlength{\itemsep}{0mm}Twitter Firehose: we randomly sampled 200
million tweets using the Twitter Firehose service. They were used
for training word embeddings and for distant supervision learning.
\item Semeval 2017 task 4 datasets of tweets, annotated according to their
general sentiment on 3 and 5 level scales; used to train the ASC model.
\item Annotated tweets from an external source\footnote{\,https://github.com/monkeylearn/sentiment-analysis-benchmark},
annotated on a 3-level scale; used to train the ASC model. 
\item Official Semeval 2018 task 1 datasets: used to train task specific
models.
\end{enumerate}
Datasets of Semeval 2017 and the external source were combined with
compression\footnote{\,Transformed 5 labels to 3: $\left\{ -2,-1\right\} \rightarrow\left\{ -1\right\} $,
$\left\{ 1,2\right\} \rightarrow\left\{ 1\right\} $, $\left\{ 0\right\} \rightarrow\left\{ 0\right\} $.}; the resulting dataset contained 88,623 tweets with the following
distribution: positive: 30097 sentences (34\%), neutral: 35818 (40\%),
negative: 22708 (26\%). Description of the official Semeval 2018 task
1 datasets can be found in \citet{SemEval2018Task1,LREC18-TweetEmo}.

\section{Preprocessing\label{sec:Data-Cleanup}}

We started by defining a cleaning pipeline that produces two cleaned
version of an original text; we refer to them as ``simple'' and
``complex'' versions. Both versions share the same initial cleaning
steps:
\begin{enumerate}
\item \setlength{\itemsep}{0mm}Word tokenization using the \href{https://stanfordnlp.github.io/CoreNLP/}{CoreNLP}
library \citep{manning-EtAl:2014:P14-5}.
\item Parts of speech (POS) tagging using the \href{http://www.cs.cmu.edu/~ark/TweetNLP/}{Tweet NLP}
tagger, trained on Twitter data \citep{owoputi2013improved}.
\item Grouping similar emojis and replacing them with representative keywords.
\item Regex: replacing URLs with a special keyword, removing duplications,
breaking \#CamelCasingHashtags into individual words. 
\end{enumerate}
The complex version contains these additional steps:
\begin{enumerate}
\item \setlength{\itemsep}{0mm}Word lemmatization, using CoreNLP. 
\item Named entity recognition (NER) using CoreNLP and replacing the entities
with representative keywords, e.g. \texttt{\_date\_}, \texttt{\_number\_},
\texttt{\_brand\_}, etc.
\item Synonym replacement, based on a manually-created dictionary. 
\item Word replacement using a Wikipedia dictionary, created by crawling
and extracting lists of places, brands and names. 
\end{enumerate}
As an example, table \ref{tab:cleaning} shows a fictitious tweet
and the results after the simple and complex cleaning stages.

\section{ASC Architecture\label{sec:RNN-Models}}

Our main contribution is an RNN network, based on GRU units with a
CNN-based attention mechanism; we will refer to it as the Amobee sentiment
classifier (ASC). It is comprised of four identical sub-models, which
differ by the input data each of them receives. Sub-model inputs are
composed of word embeddings and embeddings of the POS tags\textemdash see
section \ref{sec:Word-Embeddings} for a description of our embedding
procedure. The words were embedded in a 200 or 150 dimensional vector
spaces and the POS tags were embedded in a 8 dimensional vector space.
We pruned the tweets to have 40 words, padding shorter sentences with
a zero vector. The embeddings form the input layer. 

Next we describe the sub-model architecture; the embeddings were fed
to a bi-directional GRU layer of dimension 200. Inspired by the attention
mechanism introduced in \citet{Bahdanau:2014aa}, we extracted the
hidden states of the GRU layer; each state corresponds to a decoded
word in the GRU as it reads each tweet word by word. The hidden states
were arranged in a matrix of dimension $40\times400$ for each tweet
(bi-directionality of the GRU layer contributes a factor of 2). We
fed the hidden states to a CNN layer, instead of a weighted sum as
in the original paper. We used 6 filter sizes $[1,2,3,4,5,6]$, with
100 filters for each size. After a max-pooling layer we concatenated
all outputs, creating a 600 dimensional vector. Next was a fully connected
layer of size 30 with $\tanh$ activation, and finally a fully connected
layer of size 3 with a softmax activation function. 

\begin{figure*}[t]
\begin{centering}
\centerline{\includegraphics[bb=5bp 0bp 1870bp 600bp,clip,scale=0.245]{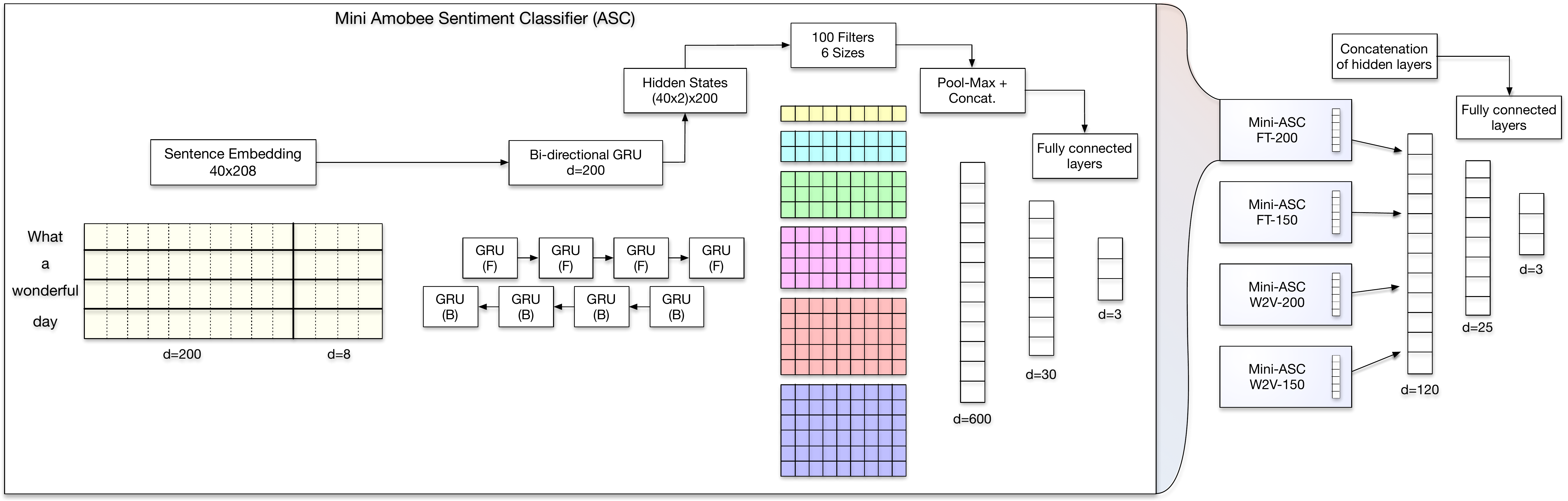}}
\par\end{centering}
\caption{\label{fig:phoenix}Architecture of the ASC nework. Each of the four
sub-models on the right has the same structure as depicted in the
central region. }
\end{figure*}

We defined 4 such sub-models with embedding inputs of the following
settings: w2v-200, w2v-150, ft-200, ft-150 (ft=FastText, w2v=Word2Vec,
see discussion in the next section). We combined the four sub-models
by extracting their hidden $d=30$ layer and concatenating them. Next
we added a fully connected $d=25$ layer with $\tanh$ activation
and a final fully connected layer of size 3. See figure \ref{fig:phoenix}
 for an illustration of the entire architecture. We used the AdaGrad
optimizer \citep{duchi2011adaptive} and a cross-entropy loss function.
We used the \href{https://keras.io/}{Keras} library \citep{chollet2015keras}
and the \href{https://www.tensorflow.org/}{TensorFlow} framework
\citep{Abadi:2016aa}.
\begin{table*}[t]
\begin{centering}
\begin{tabular}{|c|c|}
\hline 
\textbf{\footnotesize{}Original} & {\footnotesize{}@USAIRWAYS is right :-) ! Flying in September \#NiceToFly}\tabularnewline
\hline 
\textbf{\footnotesize{}Simple Cleaning} & {\footnotesize{}twitter-entity is right happy-smily ! flying in september
nice to fly}\tabularnewline
\hline 
\textbf{\footnotesize{}Complex Cleaning} & {\footnotesize{}twitter-entity be right happy-smily ! fly in \_date\_
pleasant to fly}\tabularnewline
\hline 
\end{tabular}
\par\end{centering}
\caption{\label{tab:cleaning}An example of a tweet processing, producing two
cleaned versions.}
\end{table*}

\section{Embeddings Training\label{sec:Word-Embeddings}}

Word embedding is a family of techniques in which words are encoded
as real-valued vectors of lower dimensionality. These word representations
have been used successfully in sentiment analysis tasks in recent
years. Among the popular algorithms are Word2Vec \citep{NIPS2013_5021}
and FastText \citep{bojanowski2016enriching}.

Word embeddings are useful representations of words and can uncover
hidden relationships. However, one disadvantage they have is the typical
lack of sentiment information. For example, the word vector ``good''
can be very close to the word vector ``bad'' in some trained, off-the-shelf
word embeddings. Our goal was to train word embeddings based on Twitter
data and then re-learn them so they will contain emotion-specific
sentiment.

We started with our 200 million tweets dataset; we cleaned them using
the pre-processing pipeline (described in section \ref{sec:Data-Cleanup})
and then trained generic embeddings using the \href{https://radimrehurek.com/gensim/index.html}{Gensim}
package \citep{rehurek_lrec}; we created four embeddings for the
words and two embeddings for the POS tags: for each sentence we created
a list of corresponding POS tags (there are 25 tags offered by the
tagger we used); treating the tags as words, we trained $d=8$ embeddings
using the word2vec algorithm on the simple and complex cleaned datasets.
The embeddings parameters are specified in table 
\begin{table}[b]
\begin{centering}
{\footnotesize{}}%
\begin{tabular}{|c|c|c|c|}
\cline{2-4} 
\multicolumn{1}{c|}{} & \textbf{\footnotesize{}Algorithm} & \textbf{\footnotesize{}Dimension} & \textbf{\footnotesize{}Dataset}\tabularnewline
\hline 
\multirow{4}{*}{\begin{turn}{90}
{\footnotesize{}Words}
\end{turn}} & {\footnotesize{}Word2Vec} & {\footnotesize{}200} & {\footnotesize{}Simple}\tabularnewline
\cline{2-4} 
 & {\footnotesize{}Word2Vec} & {\footnotesize{}150} & {\footnotesize{}Complex}\tabularnewline
\cline{2-4} 
 & {\footnotesize{}FastText} & {\footnotesize{}200} & {\footnotesize{}Simple}\tabularnewline
\cline{2-4} 
 & {\footnotesize{}FastText} & {\footnotesize{}150} & {\footnotesize{}Complex}\tabularnewline
\hline 
\hline 
\multirow{2}{*}{\begin{turn}{90}
{\footnotesize{}Tags}
\end{turn}} & {\footnotesize{}Word2Vec} & {\footnotesize{}200} & {\footnotesize{}Simple}\tabularnewline
\cline{2-4} 
 & {\footnotesize{}Word2Vec} & {\footnotesize{}150} & {\footnotesize{}Complex}\tabularnewline
\hline 
\end{tabular}
\par\end{centering}{\footnotesize \par}
\caption{Parameters for the word and POS tag embeddings.\label{tab:embedding_params}}
\end{table}
\ref{tab:embedding_params}.

Following \citet{tang2014learning,cliche2017bb_twtr}, who explored
training word embeddings for sentiment classification, we employed
a similar approach. We created distant supervision datasets, first,
by manually compiling 4 lists of representative words for each emotion:
anger, fear, joy and sadness; then, we built two datasets for each
emotion: the first containing tweets with the representative words
and the second does not. Each list contained about 40 words and each
dataset contained roughly 2 million tweets. We used the ASC sub-model
architecture (section \ref{sec:RNN-Models}) to train as following:
training for one epoch with embeddings set to be untrainable (fixed).
Then train for 6 epochs where the embeddings can change. 

Overall we trained 16 word embeddings\textemdash 4 embedding configurations
for each emotion. In addition, we decided to use the trained models'
final hidden layer ($d=15$) as a feature vector in the task-specific
architectures; our motivation was using them as emotion and intensity
classifiers via transfer learning. 

\section{Features Description\label{sec:Features-Extraction}}

\begin{singlespace}
In addition to our ASC models, we extracted semantic and syntactic
features, based on domain knowledge:
\end{singlespace}
\begin{itemize}
\begin{singlespace}
\item \setlength{\itemsep}{0mm}Number of magnifier and diminisher words,
e.g. ``incredibly'', ``hardly'' in each tweet.
\item Logarithm of length of sentences. 
\item Existence of elongated words, e.g. ``wowww''.
\item Fully capitalized words.
\item The symbols \#,@ appearing in the sentence. 
\item Predictions of external packages: \href{http://www.nltk.org/api/nltk.sentiment.html\#module-nltk.sentiment.vader}{Vader}
(part of the NLTK library, \citealp{Hutto:2014aa}) and \href{http://textblob.readthedocs.io/en/dev/index.html}{TextBlob}
\citep{loria2014textblob}. 
\end{singlespace}
\end{itemize}
\begin{singlespace}
Additionally, we compiled a list of 338 emojis and words in 16 categories
of emotion, annotated with scores from the set $\{0.5,1,1.5,2\}$.
For each sentence, we summed up the scores in each category, up to
a maximum value of 5, generating 16 features. The categories are:
anger, disappointed, fear, hopeful, joy, lonely, love, negative, neutral,
positive, sadness and surprise. Finally, we used the NRC Affect Intensity
lexicon \citep{mohammad2017word} containing 5814 entries; each entry
is a word with a score between 0 and 1 for a given emotion out of
the following: anger, fear, joy and sadness. We used the lexicon to
produce 4 emotion features from hashtags in the tweets; each feature
contained the largest score of all the hashtags in the tweet. For
a summary of all features used, see table \ref{tab:Features-1} in
the appendix. 
\end{singlespace}

\section{Experiments\label{sec:Experiments}}

\begin{singlespace}
Our general workflow for the tasks is as follows: for each sub-task,
we started by cleaning the datasets, obtaining two cleaned versions.
We ran a pipeline that produced all the features we designed: the
ASC predictions and the features described in section \ref{sec:Features-Extraction}.
We removed sparse features (less than 8 samples). Next, we defined
a shallow neural network with a soft-voting ensemble. We chose the
best features and meta-parameters\textemdash such as learning rate,
batch size and number of epochs\textemdash based on the dev dataset.
Finally, we generated predictions for the regression tasks. For the
classification tasks, we used a grid search method on the regression
predictions to optimize the loss. Most model trainings were conducted
on a local machine equipped with a Nvidia GTX 1080 Ti GPU. Our official
results are summarized in table 
\begin{table}
\begin{centering}
\begin{tabular}{|l|c|c|c|}
\hline 
\textbf{\footnotesize{}Task} & \textbf{\footnotesize{}Metric} & \textbf{\footnotesize{}Score} & \textbf{\footnotesize{}Ranking}\tabularnewline
\hline 
\hline 
{\footnotesize{}V-oc-Spanish} & \multirow{5}{*}{{\footnotesize{}Pearson}} & {\footnotesize{}0.765} & {\footnotesize{}1/14}\tabularnewline
\cline{1-1} \cline{3-4} 
{\footnotesize{}V-reg-Spanish} &  & {\footnotesize{}0.770} & {\footnotesize{}2/14}\tabularnewline
\cline{1-1} \cline{3-4} 
{\footnotesize{}V-oc} &  & {\footnotesize{}0.813} & {\footnotesize{}3/37}\tabularnewline
\cline{1-1} \cline{3-4} 
{\footnotesize{}EI-oc Average} &  & {\footnotesize{}0.646} & {\footnotesize{}4/39}\tabularnewline
\cline{1-1} \cline{3-4} 
{\footnotesize{}V-reg} &  & {\footnotesize{}0.843} & {\footnotesize{}5/38}\tabularnewline
\hline 
{\footnotesize{}E-c} & {\footnotesize{}Jaccard} & {\footnotesize{}0.566} & {\footnotesize{}6/35}\tabularnewline
\hline 
{\footnotesize{}EI-reg Average} & {\footnotesize{}Pearson} & {\footnotesize{}0.721} & {\footnotesize{}13/48}\tabularnewline
\hline 
\end{tabular}
\par\end{centering}
\caption{Summary of results.\label{tab:Results}}
\end{table}
\ref{tab:Results}.
\end{singlespace}

\subsection{Valence Prediction\label{subsec:Valence}}

In the valence sub-tasks, we identified how intense a general sentiment
(valence) is; the score is either in a continuous scale between 0
and 1 or classified into 7 ordinal classes $\{-3,-2,-1,0,1,2,3\}$,
and is evaluated using the Pearson correlation coefficient. 

We started with the regression task and defined the following model:
first, we normalized the features to have zero mean and $\text{SD}=1$.
Then, we inserted 300 instances of fully connected layers of size
3, with a softmax activation and no bias term. For each copy, we applied
the function $f(x)=\left(x_{0}-x_{2}\right)/2+0.5$ where $x_{0},x_{2}$
are the 1st and 3rd component of each hidden layer. Our aim was transforming
the label predictions of the ASCs (trained on 3-label based sentiment
annotation) into a regression score such that high certainty in either
label (negative, neutral or positive) would produce scores close to
0, 0.5 or 1, respectively. Finally, we calculated the mean of all
300 prediction to get the final node; this is also known as a soft-voting
ensemble. We used the Adam optimizer \citep{kingma2014adam} with
default values, mean-square-error loss function, batch size of 400
and 65 epochs of training. For an illustration of the network, see
figure 
\begin{figure}[t]
\begin{centering}
\includegraphics[scale=0.45]{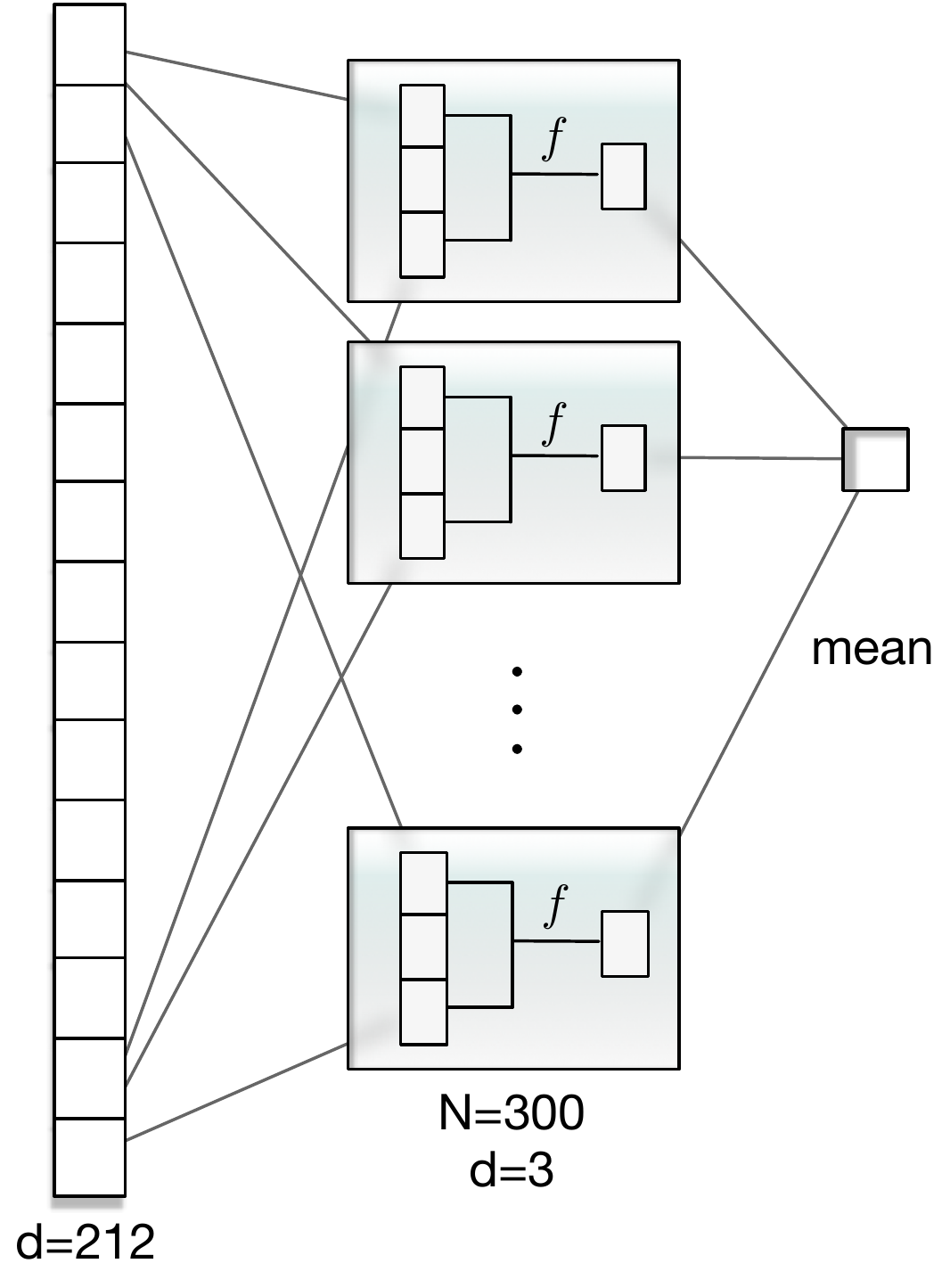}
\par\end{centering}
\caption{Architecture of the final classifier in the valence sub-tasks, where
$f=\left(x_{0}-x_{2}\right)/2+0.5$ and the input dimension is 212
for the V-reg sub-task. \label{fig:4-emotion}}
\end{figure}
\ref{fig:4-emotion}. We experimented with the dev dataset, testing
different subsets of the features. Finally we produced predictions
for the regression sub-task V-reg.

We analyzed the relative contribution of each feature by measuring
variable importance using \citet{pratt1987dividing} approach. We
calculated scores $d_{i}$ for each feature using the following formula:
$d_{i}=\hat{\beta}_{i}\,\hat{\rho}_{i}/R^{2}$ where $\hat{\beta}_{i}$
denotes the sample estimation of the feature, $\hat{\rho}_{i}$ is
the simple correlation between the labels and the $i$th feature and
$R^{2}$ is the coefficient of determination (see \citealt{thomas1998variable}).
We present the relative contribution of each feature in figure 
\begin{figure*}
\begin{centering}
\centerline{\includegraphics[scale=0.425]{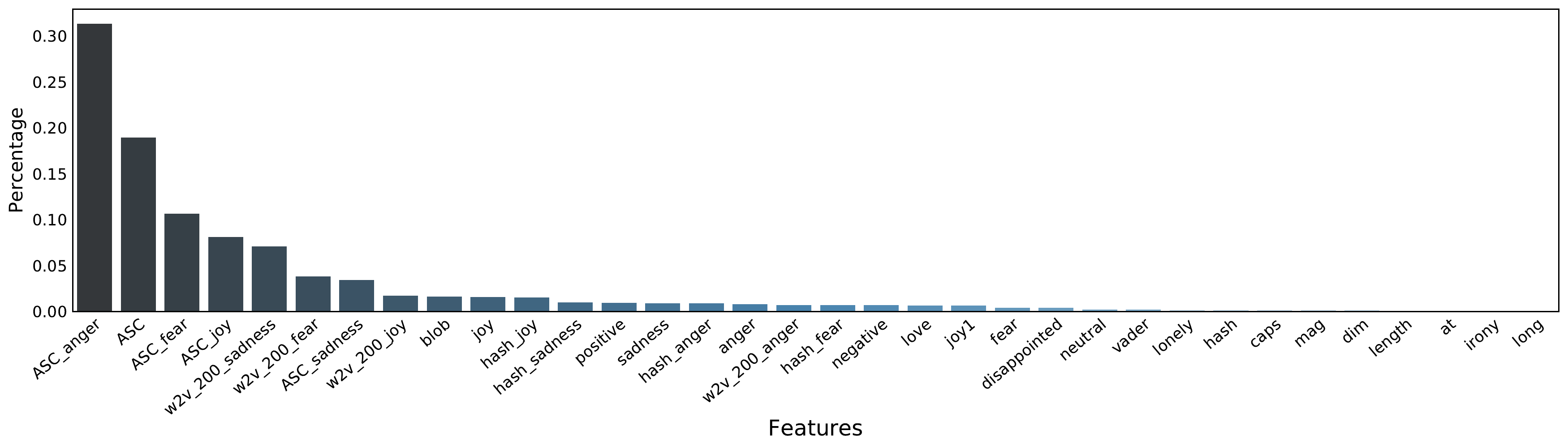}}
\par\end{centering}
\caption{Relative contribution of features in the valence regression sub-task.
\label{fig:Weights-1}}
\end{figure*}
\ref{fig:Weights-1} and the top 10 features in table 
\begin{table}[b]
\begin{centering}
\begin{tabular}{|l|c|c|}
\hline 
\textbf{\footnotesize{}Name} & \textbf{\footnotesize{}Dim.} & \textbf{\footnotesize{}\%}\tabularnewline
\hline 
\hline 
\textbf{\scriptsize{}ASC\_anger} & {\footnotesize{}25} & {\footnotesize{}31.38\%}\tabularnewline
\hline 
\textbf{\scriptsize{}ASC} & {\footnotesize{}25} & {\footnotesize{}18.92\%}\tabularnewline
\hline 
\textbf{\scriptsize{}ASC\_fear} & {\footnotesize{}25} & {\footnotesize{}10.63\%}\tabularnewline
\hline 
\textbf{\scriptsize{}ASC\_joy} & {\footnotesize{}25} & {\footnotesize{}8.13\%}\tabularnewline
\hline 
\textbf{\scriptsize{}W2V\_200\_sadness} & {\footnotesize{}15} & {\footnotesize{}7.10\%}\tabularnewline
\hline 
\textbf{\scriptsize{}W2V\_200\_fear} & {\footnotesize{}15} & {\footnotesize{}3.82\%}\tabularnewline
\hline 
\textbf{\scriptsize{}ASC\_sadness} & {\footnotesize{}25} & {\footnotesize{}3.46\%}\tabularnewline
\hline 
\textbf{\scriptsize{}W2V\_200\_joy} & {\footnotesize{}15} & {\footnotesize{}1.74\%}\tabularnewline
\hline 
\textbf{\scriptsize{}Blob} & {\footnotesize{}1} & {\footnotesize{}1.64\%}\tabularnewline
\hline 
\textbf{\scriptsize{}Joy} & {\footnotesize{}1} & {\footnotesize{}1.60\%}\tabularnewline
\hline 
\end{tabular}\caption{Relative contribution of features in the valence regression sub-task.\label{tab:features-valence}}
\par\end{centering}
\end{table}
\ref{tab:features-valence}. We can see that the ASC models, both
general and emotion-specific, contributed about 72\% of the total
contribution made by all features, in this sub-task.

For the ordinal classification task, we used the predictions of the
regression task on the sentences, which were the same in both tasks.
Using a grid search method, we partitioned the regression scores into
7 categories such that the Pearson correlation coefficient was maximized.
We submitted the classes predictions as sub-task V-oc. Our final scores
were 0.843, 0.813 in the regression and classification sub-tasks,
respectively. 

\subsection{Emotion Intensity\label{subsec:Emotion-Intensity}}

In the emotion intensity sub-tasks, we identified how intense a given
emotion is in the given tweets. The four emotions were: anger, fear,
joy and sadness; the score is either in a scale between 0 and 1 or
classified into 4 ordinal classes $\{0,1,2,3\}$. Performance was
evaluated using the Pearson correlation coefficient. Our approach
was similar to the valence tasks; first we generated features, then
we used the same architecture as in the valence sub-tasks, depicted
in figure \ref{fig:4-emotion}. However, in these sub-tasks we used
the emotion-specific embeddings for each emotion sub-task. We generated
regression predictions and submitted them as the EI-reg sub-tasks;
finally we carried a grid search for the best partition, maximizing
the Pearson correlation and submitted the classes predictions as sub-tasks
EI-oc. For a summary of the training parameters used in the regression
sub-tasks, see table 
\begin{table}
\begin{centering}
\begin{tabular}[t]{|c|c|c|c|c|}
\hline 
\textbf{\footnotesize{}EI-reg} & \textbf{\footnotesize{}Anger} & \textbf{\footnotesize{}Fear} & \textbf{\footnotesize{}Joy} & \textbf{\footnotesize{}Sadness}\tabularnewline
\hline 
\hline 
{\footnotesize{}Features} & {\footnotesize{}204} & {\footnotesize{}274} & {\footnotesize{}150} & {\footnotesize{}181}\tabularnewline
\hline 
{\footnotesize{}Learning rate} & {\footnotesize{}$10^{-4}$} & {\footnotesize{}$10^{-5}$} & {\footnotesize{}$10^{-5}$} & {\footnotesize{}$3\cdot10^{-5}$}\tabularnewline
\hline 
{\footnotesize{}Epochs} & {\footnotesize{}330} & {\footnotesize{}700} & {\footnotesize{}700} & {\footnotesize{}1000}\tabularnewline
\hline 
\end{tabular}
\par\end{centering}
\caption{Summary of training parameters for the emotion intensity regression
tasks. \label{tab:4-emot-params}}
\end{table}
\ref{tab:4-emot-params}. 

Our system performed as following: in the regression tasks, the scores
were: 0.748, 0.670, 0.748, 0.721 for the anger, fear, joy and sadness,
respectively, with a macro-average of 0.721. In the classification
tasks, the scores were: 0.667, 0.536, 0.705, 0.673 for the anger,
fear, joy and sadness, respectively, with a macro-average of 0.646. 

\subsection{Multi-label Classification\label{subsec:Multi-label}}

In the multi-label classification sub-task, we had to label tweets
with respect to 11 emotions: anger, anticipation, disgust, fear, joy,
love, optimism, pessimism, sadness, surprise and trust. The score
was evaluated using the Jaccard similarity coefficient. We started
with the same cleaning and feature-generation pipelines as before,
creating an input layer of size 217. We added a fully connected layer
of size 100 with $\tanh$ activation. Next there were 300 instances
of fully connected layers of size 11 with sigmoid activation function.
We calculated the mean of all $d=11$ vectors, producing the final
$d=11$ vector. For an illustration, see figure 
\begin{figure}
\begin{centering}
\includegraphics[scale=0.32]{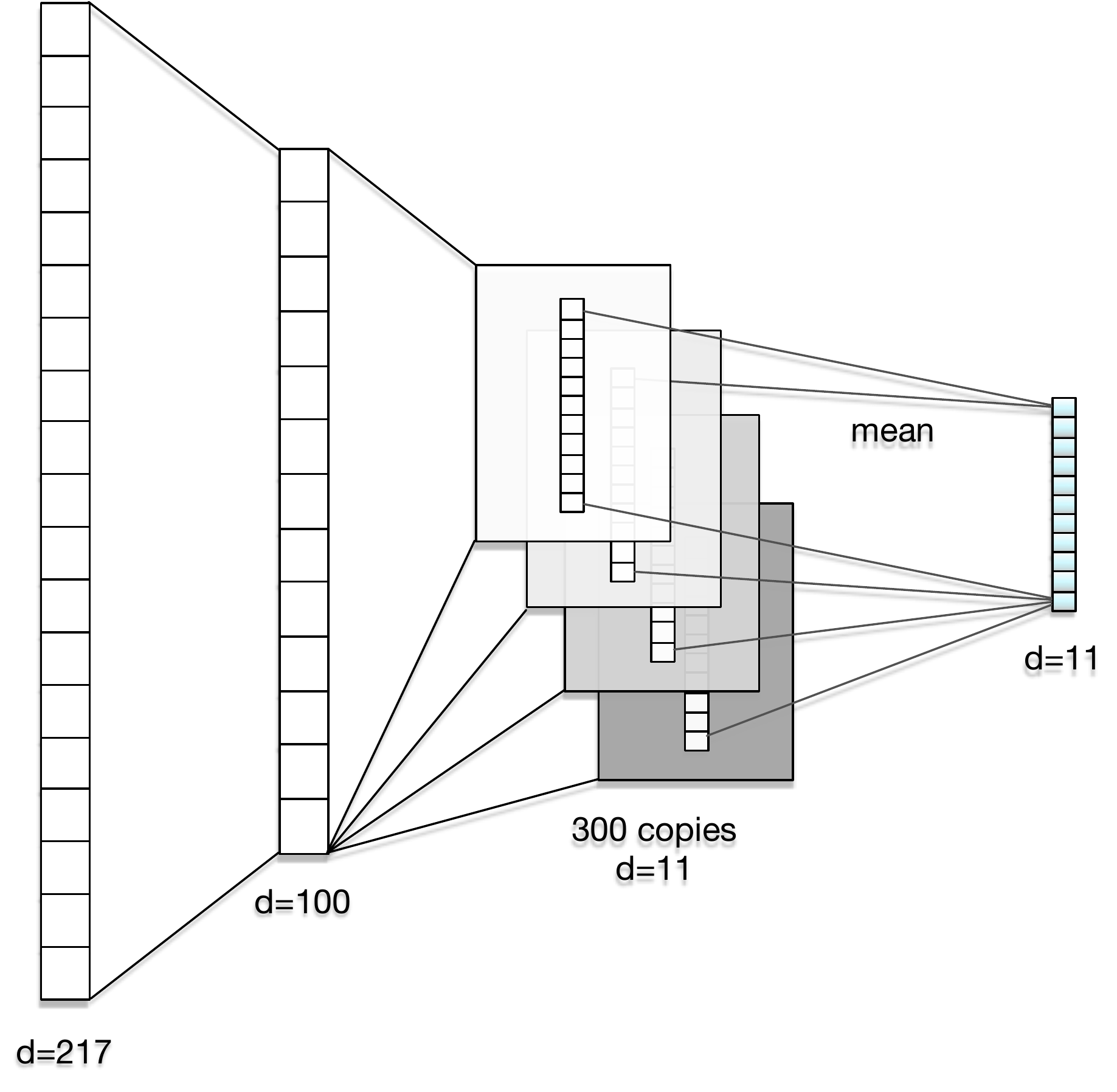}
\par\end{centering}
\caption{Architecture of the multi-label sub-task E-c.\label{fig:Multi}}
\end{figure}
\ref{fig:Multi} for an illustration. We used the following loss function,
based on Tanimoto distance: $L(y,\tilde{y})=1-\frac{y\cdot\tilde{y}}{\|y+\tilde{y}\|_{1}-y\cdot\tilde{y}+\epsilon},$
where $\|\cdot\|_{1}$ is an $L^{1}$ norm and $\epsilon=10^{-7}$
is used for numerical stability. We trained with a batch size of 10,
for 40 epochs with Adam optimization with default parameters. Our
final score was 0.566. 

\subsection{Spanish Valence Tasks\label{subsec:Spanish}}

We participated in the Spanish valence tasks to examine the current
state of neural machine translation (NMT) algorithms. We used the
Google \href{https://cloud.google.com/translate/}{Cloud Translation API}
to translate the Spanish training, development and test datasets for
the two valence tasks from Spanish to English. We then treated the
tasks the same way as the English valence tasks, using the same cleaning
and feature extraction pipelines and the same architecture described
in section \ref{subsec:Valence} to generate regression and classification
predictions. We reached 1st and 2nd places in the classification and
regression sub-tasks, with scores of 0.765, 0.770, respectively. 

\section{Review and Conclusions\label{sec:Review-and-Conclusions}}

In this paper we described the system developed to participate in
the Semeval 2018 task 1 workshop. We reached 3rd place in the valence
ordinal classification sub-task and 5th place in the valence regression
sub-task. In the Spanish valence tasks, we reached 1st and 2nd places
in the classification and regression sub-tasks, respectively. In the
emotions intensity sub-tasks we reached 4th and 13th places in the
classification and regression sub-tasks, respectively. 

Summarizing the methods used: training of word embeddings based on
a Twitter corpus (200M tweets), developing and using Amobee sentiment
classifier (ASC) architecture\textemdash a bi-directional GRU layer
with a CNN-based attention mechanism and an additional hidden layer\textemdash used
to adjust the embeddings to include emotional context, and finally
a shallow feed-forward NN with a stack-based ensemble of final hidden
layers from all previous classifiers we trained. This form of transfer
learning proved to be important, as the hidden layers features achieved
a significant contribution to minimizing the loss.

Overall, we had better performance in the valence tasks, both in English
and Spanish. We posit this is due to the fact our annotated supervised
training dataset (non task-specific) was based on Semeval 2017 task
4, which focused on valence classification. In addition, the annotations
in Semeval 2017 were label-based, lending themselves more easily to
the ordinal classification tasks. In the Spanish tasks, we used external
translation (Google API) and achieved good results without the use
of Spanish-specific features. 

\subsection*{Acknowledgment}

We thank Zohar Kelrich for assisting in translating the Spanish datasets
to English. 

\bibliographystyle{acl_natbib}
\bibliography{semeval2018}

\begin{thebibliography}{20}
\expandafter\ifx\csname natexlab\endcsname\relax\def\natexlab#1{#1}\fi

\bibitem[{Abadi et~al.(2016)Abadi, Barham, Chen, Chen, Davis, Dean, Devin,
  Ghemawat, Irving, Isard et~al.}]{Abadi:2016aa}
Mart{\'\i}n Abadi, Paul Barham, Jianmin Chen, Zhifeng Chen, Andy Davis, Jeffrey
  Dean, Matthieu Devin, Sanjay Ghemawat, Geoffrey Irving, Michael Isard, et~al.
  2016.
\newblock Tensorflow: A system for large-scale machine learning.
\newblock In \emph{OSDI}, volume~16, pages 265--283.

\bibitem[{Bahdanau et~al.(2014)Bahdanau, Cho, and Bengio}]{Bahdanau:2014aa}
Dzmitry Bahdanau, Kyunghyun Cho, and Yoshua Bengio. 2014.
\newblock \href {http://arxiv.org/abs/1409.0473} {Neural machine translation by
  jointly learning to align and translate}.
\newblock \emph{CoRR}, abs/1409.0473.

\bibitem[{Bojanowski et~al.(2016)Bojanowski, Grave, Joulin, and
  Mikolov}]{bojanowski2016enriching}
Piotr Bojanowski, Edouard Grave, Armand Joulin, and Tomas Mikolov. 2016.
\newblock Enriching word vectors with subword information.
\newblock \emph{arXiv preprint arXiv:1607.04606}.

\bibitem[{Cho et~al.(2014)Cho, van Merrienboer, Bahdanau, and
  Bengio}]{Cho:2014aa}
KyungHyun Cho, Bart van Merrienboer, Dzmitry Bahdanau, and Yoshua Bengio. 2014.
\newblock \href {http://arxiv.org/abs/1409.1259} {On the properties of neural
  machine translation: Encoder-decoder approaches}.
\newblock \emph{CoRR}, abs/1409.1259.

\bibitem[{Chollet et~al.(2015)}]{chollet2015keras}
Fran\c{c}ois Chollet et~al. 2015.
\newblock Keras.
\newblock \url{https://github.com/keras-team/keras}.

\bibitem[{Cliche(2017)}]{cliche2017bb_twtr}
Mathieu Cliche. 2017.
\newblock Bb\_twtr at semeval-2017 task 4: Twitter sentiment analysis with cnns
  and lstms.
\newblock \emph{arXiv preprint arXiv:1704.06125}.

\bibitem[{Duchi et~al.(2011)Duchi, Hazan, and Singer}]{duchi2011adaptive}
John Duchi, Elad Hazan, and Yoram Singer. 2011.
\newblock Adaptive subgradient methods for online learning and stochastic
  optimization.
\newblock \emph{Journal of Machine Learning Research}, 12(Jul):2121--2159.

\bibitem[{Hutto and Gilbert(2014)}]{Hutto:2014aa}
C.J. Hutto and E.E. Gilbert. 2014.
\newblock Vader: A parsimonious rule-based model for sentiment analysis of
  social media text.
\newblock In \emph{Eighth International Conference on Weblogs and Social Media
  (ICWSM-14)}, Ann Arbor, MI.

\bibitem[{Kingma and Ba(2014)}]{kingma2014adam}
Diederik~P Kingma and Jimmy Ba. 2014.
\newblock Adam: A method for stochastic optimization.
\newblock \emph{arXiv preprint arXiv:1412.6980}.

\bibitem[{Loria et~al.(2014)Loria, Keen, Honnibal, Yankovsky, Karesh, Dempsey
  et~al.}]{loria2014textblob}
Steven Loria, P~Keen, M~Honnibal, R~Yankovsky, D~Karesh, E~Dempsey, et~al.
  2014.
\newblock Textblob: simplified text processing.
\newblock \emph{Secondary TextBlob: Simplified Text Processing}.

\bibitem[{Manning et~al.(2014)Manning, Surdeanu, Bauer, Finkel, Bethard, and
  McClosky}]{manning-EtAl:2014:P14-5}
Christopher~D. Manning, Mihai Surdeanu, John Bauer, Jenny Finkel, Steven~J.
  Bethard, and David McClosky. 2014.
\newblock \href {http://www.aclweb.org/anthology/P/P14/P14-5010} {The
  {Stanford} {CoreNLP} natural language processing toolkit}.
\newblock In \emph{Association for Computational Linguistics (ACL) System
  Demonstrations}, pages 55--60.

\bibitem[{Mikolov et~al.(2013)Mikolov, Sutskever, Chen, Corrado, and
  Dean}]{NIPS2013_5021}
Tomas Mikolov, Ilya Sutskever, Kai Chen, Greg~S Corrado, and Jeff Dean. 2013.
\newblock \href
  {http://papers.nips.cc/paper/5021-distributed-representations-of-words-and-phrases-and-their-compositionality.pdf}
  {Distributed representations of words and phrases and their
  compositionality}.
\newblock In C.~J.~C. Burges, L.~Bottou, M.~Welling, Z.~Ghahramani, and K.~Q.
  Weinberger, editors, \emph{Advances in Neural Information Processing Systems
  26}, pages 3111--3119. Curran Associates, Inc.

\bibitem[{Mohammad(2017)}]{mohammad2017word}
Saif~M Mohammad. 2017.
\newblock Word affect intensities.
\newblock \emph{arXiv preprint arXiv:1704.08798}.

\bibitem[{Mohammad et~al.(2018)Mohammad, Bravo-Marquez, Salameh, and
  Kiritchenko}]{SemEval2018Task1}
Saif~M. Mohammad, Felipe Bravo-Marquez, Mohammad Salameh, and Svetlana
  Kiritchenko. 2018.
\newblock Semeval-2018 {T}ask 1: {A}ffect in tweets.
\newblock In \emph{Proceedings of International Workshop on Semantic Evaluation
  (SemEval-2018)}, New Orleans, LA, USA.

\bibitem[{Mohammad and Kiritchenko(2018)}]{LREC18-TweetEmo}
Saif~M. Mohammad and Svetlana Kiritchenko. 2018.
\newblock Understanding emotions: A dataset of tweets to study interactions
  between affect categories.
\newblock In \emph{Proceedings of the 11th Edition of the Language Resources
  and Evaluation Conference}, Miyazaki, Japan.

\bibitem[{Owoputi et~al.(2013)Owoputi, O'Connor, Dyer, Gimpel, Schneider, and
  Smith}]{owoputi2013improved}
Olutobi Owoputi, Brendan O'Connor, Chris Dyer, Kevin Gimpel, Nathan Schneider,
  and Noah~A Smith. 2013.
\newblock Improved part-of-speech tagging for online conversational text with
  word clusters.
\newblock Association for Computational Linguistics.

\bibitem[{Pratt(1987)}]{pratt1987dividing}
John~W Pratt. 1987.
\newblock Dividing the indivisible: Using simple symmetry to partition variance
  explained.
\newblock In \emph{Proceedings of the second international Tampere conference
  in statistics, 1987}, pages 245--260. Department of Mathematical Sciences,
  University of Tampere.

\bibitem[{{\v R}eh{\r u}{\v r}ek and Sojka(2010)}]{rehurek_lrec}
Radim {\v R}eh{\r u}{\v r}ek and Petr Sojka. 2010.
\newblock {Software Framework for Topic Modelling with Large Corpora}.
\newblock In \emph{{Proceedings of the LREC 2010 Workshop on New Challenges for
  NLP Frameworks}}, pages 45--50, Valletta, Malta. ELRA.
\newblock \url{http://is.muni.cz/publication/884893/en}.

\bibitem[{Tang et~al.(2014)Tang, Wei, Yang, Zhou, Liu, and
  Qin}]{tang2014learning}
Duyu Tang, Furu Wei, Nan Yang, Ming Zhou, Ting Liu, and Bing Qin. 2014.
\newblock Learning sentiment-specific word embedding for twitter sentiment
  classification.
\newblock In \emph{Proceedings of the 52nd Annual Meeting of the Association
  for Computational Linguistics (Volume 1: Long Papers)}, volume~1, pages
  1555--1565.

\bibitem[{Thomas et~al.(1998)Thomas, Hughes, and Zumbo}]{thomas1998variable}
D~Roland Thomas, Edward Hughes, and Bruno~D Zumbo. 1998.
\newblock On variable importance in linear regression.
\newblock \emph{Social Indicators Research}, 45(1-3):253--275.

\end{thebibliography}

\appendix
\onecolumn

\section{Features List}

List of features used as inputs for the task-specific models. 

\begin{table}[h]
\begin{centering}
\begin{tabular}{|>{\raggedright}p{0.3\paperwidth}|>{\raggedright}p{0.28\paperwidth}|l|}
\hline 
\textbf{\footnotesize{}Name} & \textbf{\footnotesize{}Description} & \textbf{\footnotesize{}Dim.}\tabularnewline
\hline 
{\footnotesize{}ASC} & {\footnotesize{}ASC model hidden layer.} & {\footnotesize{}25}\tabularnewline
\hline 
{\footnotesize{}ASC x \{anger,fear,joy,sadness\}} & {\footnotesize{}Emotion specific ASC hidden layers. } & {\footnotesize{}$4\times25$}\tabularnewline
\hline 
{\footnotesize{}at} & {\footnotesize{}`@' symbol in tweet.} & {\footnotesize{}1}\tabularnewline
\hline 
{\footnotesize{}blob} & {\footnotesize{}TextBlob sentiment library.} & {\footnotesize{}1}\tabularnewline
\hline 
{\footnotesize{}caps} & {\footnotesize{}Occurrence of all capitalized words.} & {\footnotesize{}1}\tabularnewline
\hline 
{\footnotesize{}dim} & {\footnotesize{}Diminisher words.} & {\footnotesize{}1}\tabularnewline
\hline 
{\footnotesize{}\{ft,w2v\} x \{150,200\} x \{anger,fear,joy,sadness\}} & {\footnotesize{}Hidden layers of models used to re-train the embeddings.} & {\footnotesize{}$4\times4\times15$}\tabularnewline
\hline 
{\footnotesize{}hash} & {\footnotesize{}`\#' symbol in tweet.} & {\footnotesize{}1}\tabularnewline
\hline 
{\footnotesize{}hash x \{anger,fear,joy,sadness\}} & {\footnotesize{}Affection lexicon of hashtags.} & {\footnotesize{}4}\tabularnewline
\hline 
{\footnotesize{}irony} & {\footnotesize{}Occurrence of \#irony or \#sarcasm hashtags.} & {\footnotesize{}1}\tabularnewline
\hline 
{\footnotesize{}length} & {\footnotesize{}Logarithm of sentence length.} & {\footnotesize{}1}\tabularnewline
\hline 
{\footnotesize{}long} & {\footnotesize{}Elongated words, `wowwww'.} & {\footnotesize{}1}\tabularnewline
\hline 
{\footnotesize{}mag} & {\footnotesize{}Magnifiers.} & {\footnotesize{}1}\tabularnewline
\hline 
{\footnotesize{}vader} & {\footnotesize{}Vader sentiment library.} & {\footnotesize{}3}\tabularnewline
\hline 
\multicolumn{1}{>{\raggedright}p{0.3\paperwidth}}{} & \multicolumn{1}{>{\raggedright}p{0.28\paperwidth}}{} & \multicolumn{1}{l}{}\tabularnewline
\hline 
{\footnotesize{}negative} & {\footnotesize{}Negative emojis.} & {\footnotesize{}1}\tabularnewline
\hline 
{\footnotesize{}neutral} & {\footnotesize{}Neutral emojis.} & {\footnotesize{}1}\tabularnewline
\hline 
{\footnotesize{}positive} & {\footnotesize{}Positive emojis.} & {\footnotesize{}1}\tabularnewline
\hline 
{\footnotesize{}anger/1} & \multirow{9}{0.28\paperwidth}{{\footnotesize{}Detection of emojis and words related to the given
emotion, taken from a manually annotated list.}} & {\footnotesize{}2}\tabularnewline
\cline{1-1} \cline{3-3} 
{\footnotesize{}fear/1} &  & {\footnotesize{}2}\tabularnewline
\cline{1-1} \cline{3-3} 
{\footnotesize{}joy/1} &  & {\footnotesize{}2}\tabularnewline
\cline{1-1} \cline{3-3} 
{\footnotesize{}sadness/1} &  & {\footnotesize{}2}\tabularnewline
\cline{1-1} \cline{3-3} 
{\footnotesize{}love} &  & {\footnotesize{}1}\tabularnewline
\cline{1-1} \cline{3-3} 
{\footnotesize{}surprise} &  & {\footnotesize{}1}\tabularnewline
\cline{1-1} \cline{3-3} 
{\footnotesize{}disappointed} &  & {\footnotesize{}1}\tabularnewline
\cline{1-1} \cline{3-3} 
{\footnotesize{}lonely} &  & {\footnotesize{}1}\tabularnewline
\cline{1-1} \cline{3-3} 
{\footnotesize{}hopeful} &  & {\footnotesize{}1}\tabularnewline
\hline 
\end{tabular}
\par\end{centering}
\caption{Complete list of features generated from datasets. \label{tab:Features-1}}
\end{table}

\end{document}